\documentclass[conference]{IEEEtran}
\usepackage{times}

\usepackage[numbers]{natbib}
\usepackage{multicol}
\usepackage{graphics} % for pdf, bitmapped graphics files
\usepackage{epsfig} % for postscript graphics files
\usepackage{mathptmx} % assumes new font selection scheme installed
\usepackage{times} % assumes new font selection scheme installed
\usepackage{amsmath} % assumes amsmath package installed
\usepackage{amsfonts}
\usepackage{amsthm}
\usepackage{amssymb}  % assumes amsmath package installed
\usepackage{xcolor}

\newcommand{\Hu}{\mathcal{H}}
\newcommand{\dem}{\xi^{D}}

\newcommand{\pr}{\mathbb{P}}

\newcommand{\uH}{u_{H}}

\newcommand{\UH}{U_{H}}

\newcommand{\dyn}{f}

\newcommand{\RH}{R_{H}}
\newcommand{\B}[2]{B({#1}, {#2})}
\newcommand{\w}{\textbf{w}}
\newcommand{\norm}[1]{||{#1}||}
\newcommand{\rk}[1]{\xi_{\sigma({#1})}}
\newcommand{\rank}{r_\sigma}

\newcommand{\prg}[1]{\vspace{0.05in}\noindent\textbf{{#1}}}
\newcommand{\subprg}[1]{\vspace{0.05in}\noindent\textit{\textbf{{#1}}}}

\renewcommand{\baselinestretch}{0.96}

\begin{document}

% paper title
\title{Learning Reward Functions by \\ Integrating Human Demonstrations and Preferences}

\author{\authorblockN{Malayandi Palan*, Nicholas C. Landolfi*, Gleb Shevchuk, Dorsa Sadigh}
\authorblockA{Computer Science, Stanford University\\
Email: \{malayandi, lando, glebs, dorsa\}@stanford.edu}
}

\maketitle

\begin{abstract}
Our goal is to accurately and efficiently learn reward functions for autonomous robots. Current approaches to this problem include inverse reinforcement learning (IRL), which uses expert demonstrations, and preference-based learning, which iteratively queries the user for her preferences between trajectories. In robotics however, IRL often struggles because it is difficult to get high-quality demonstrations; conversely, preference-based learning is very inefficient since it attempts to learn a continuous, high-dimensional function from binary feedback. We propose a new framework for reward learning, DemPref, that uses \textit{both} demonstrations and preference queries to learn a reward function. Specifically, we (1) use the demonstrations to learn a coarse prior over the space of reward functions, to reduce the effective size of the space from which queries are generated; and (2) use the demonstrations to ground the (active) query generation process, to improve the quality of the generated queries. Our method alleviates the efficiency issues faced by standard preference-based learning methods and does not exclusively depend on (possibly low-quality) demonstrations. In numerical experiments, we find that DemPref is significantly more efficient than a standard active preference-based learning method. In a user study, we compare our method to a standard IRL method; we find that users rated the robot trained with DemPref as being more successful at learning their desired behavior, and preferred to use the DemPref system (over IRL) to train the robot.
\end{abstract}

\IEEEpeerreviewmaketitle

\section{Introduction}

\begin{figure}[t]
\includegraphics[width=\columnwidth]{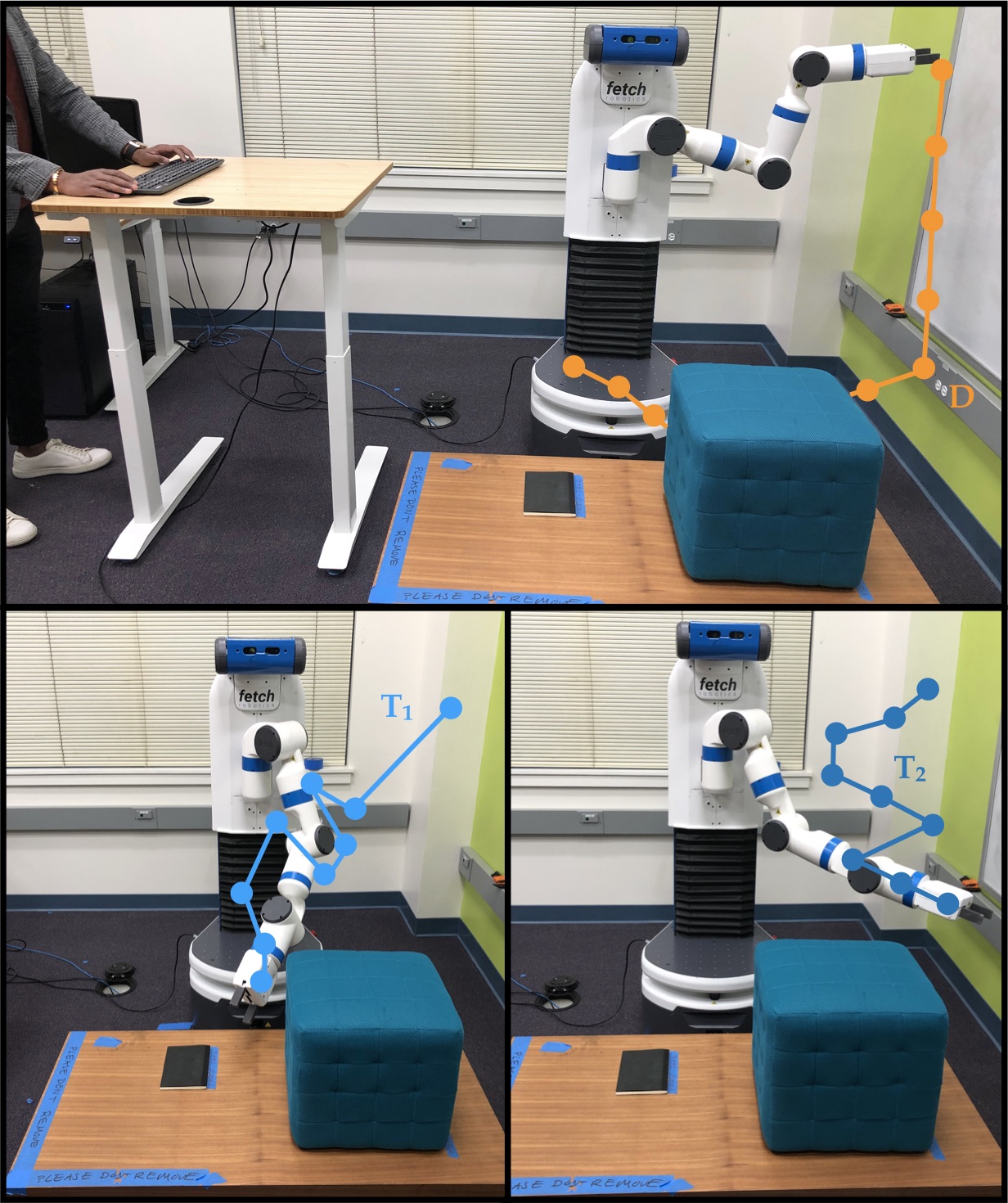}
\centering
\caption{We show how to leverage multi-modal human input -- both demonstrations and preferences -- to learn personalized human reward functions accurately and efficiently. At a high-level, our framework, DemPref, works as follows: the human provides demonstrations by tele-operating the robot (top). These demonstrations are used to generate queries, to which the human responds by selecting her preferred trajectory (bottom).}
\label{front_fig}
\vspace{-0.2in}
\end{figure}

In recent years, we have witnessed significant improvements in a robot's ability to optimize a reward function, which makes it all the more crucial that these robots are optimizing the \textit{correct} reward function. More often than not, these reward functions are hand-coded by a system designer. However, it is difficult (if not impossible) to craft a reward function that perfectly encodes the robot's desired behavior in every setting it may encounter; this can lead to unintended, and perhaps even dangerous, consequences. For example, recent work observed that a simulated robot with a simple reward function, when tasked to slide a block on a table, learned to slide the \textit{entire table} instead \cite{chopra_2018}. Similarly, it may seem reasonable to provide an autonomous vehicle with a reward function that encourages it to maintain a high speed on a highway; however, recent work observed that, with such a reward function, the (simulated) autonomous vehicle learned to spin rapidly in circles instead of driving down the road \cite{kelcey_2017}.\let\thefootnote\relax\footnotetext{*Equal contribution.}

We therefore concern ourselves with the problem of accurately learning a human's reward function without the need for a system designer to hard-code such a function. Current approaches to this problem include inverse reinforcement learning (IRL) \cite{abbeel2004apprenticeship,ramachandran2007bayesian,ziebart2008maximum}, where we learn a reward function directly from expert demonstrations of the task, and preference-based learning \cite{eric2008active,dorsa2017active}, where we learn a reward function by repeatedly asking a human to pick between two trajectories. While these methods have found some success, they still struggle in practice, especially in robotics.

IRL assumes access to high-quality demonstrations of the task. However, this is rarely available in robotics, where it is difficult to control high degree-of-freedom (DOF) robots \cite{dragan2012formalizing,javdani2015shared,khurshid2015data}. Preference-based learning methods on the other are hand are very inefficient since they attempt to learn a continuous reward function from binary feedback. Often, they require solving an optimization problem over the control space at each step \cite{biyik2018batch, dorsa2017active}, but this is challenging in robotics, where the control space is very complex and high-dimensional.

\begin{figure*}[t]
\includegraphics[width=\textwidth]{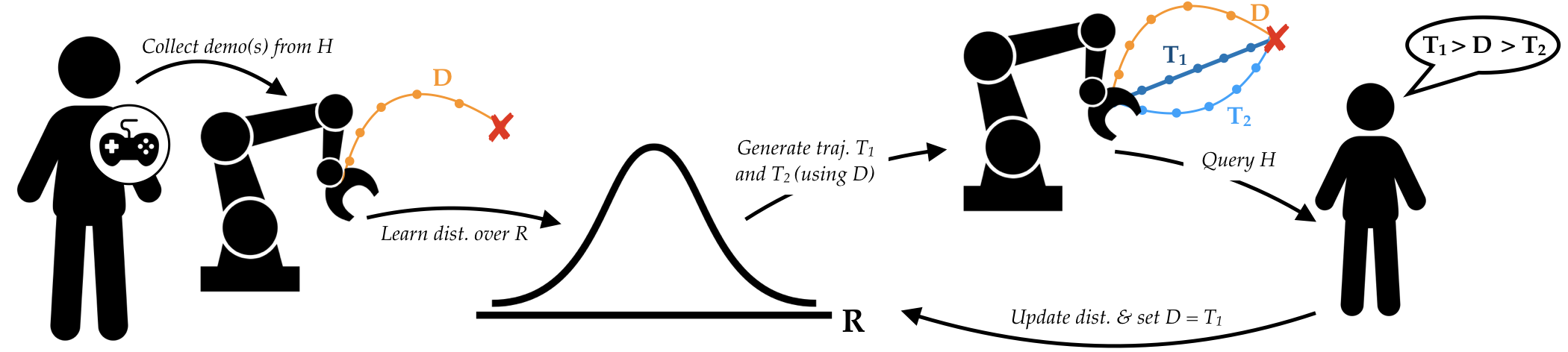}
\centering
\caption{The full DemPref framework. The human user provides demonstrations, which are used to learn a prior over reward functions. Then, we actively generate two trajectories that, \textit{together with the human's demonstration}, maximize the expected information gained from querying the human with them. We then query the human for her ranking of the trajectories. We use that information to update the distribution over reward functions and use the chosen trajectory in place of the demonstration in the next query. We repeat this process of generating trajectories and querying the human until convergence.}
\label{schem_fig}
\vspace{-0.2in}
\end{figure*}

Our key insight is that demonstrations and preferences are complementary: \emph{preferences queries are comparatively more accurate but less informative than demonstrations.} We use this insight to derive a unified, active framework for reward learning, DemPref, that leverages both modes of feedback to learn reward functions accurately and efficiently.

Specifically, we use an IRL method to learn a prior over reward functions from demonstrations, which significantly reduces the size of the space of possible reward functions. We then derive a modified active, preference-based learning method -- which leverages the demonstrations to generate better queries -- and use this method with our learned prior to find the true reward function. Our framework is (1) more robust to low-quality demonstrations than IRL since we do not solely rely on demonstrations and (2) more efficient than standard preference-based learning methods since in the preference-learning stage, our framework has access to additional structure (from the demonstrations) and needs only search over a much smaller space of reward functions. We verify these claims in a series of numerical experiments in simulation and in a user study on a physical, 7-DOF Fetch mobile manipulator robot \cite{wise2016fetch} with 15 participants. In the supplementary material, we provide videos of our framework in action along with our code, to complement the text descriptions and figures in the main paper. Concretely, our contributions are:

\prg{1. Integrated Framework for Reward Learning.} We propose a unified framework for reward learning, DemPref, that utilizes both demonstrations and preference queries to learn a personalized human reward function.

\prg{2. Method for Generating Better Preference Queries.} We utilize the additional structure of demonstrations to derive a modified active, preference-based learning algorithm for DemPref, to improve the quality of the generated queries.

\prg{3. Numerical Analysis of Algorithm.} We verify in simulation that our method can learn reward functions more efficiently than a standard preference-based learning algorithm.

\prg{4. User Study on a Physical Robot.} We conduct a user study on a 7-DOF manipulator and empirically verify  (1) that our method is more successful than IRL at learning human reward functions (as evaluated by the users themselves) and (2) that people prefer to use DemPref over IRL for training robots.
\section{Background on Reward Learning}

\prg{Inverse Reinforcement Learning.} IRL is a popular framework for learning reward functions from human demonstrations \cite{abbeel2004apprenticeship,argall2009survey,brown2019risk, cui2018active,hadfield2017inverse,levine2012continuous,malik2018efficient,ramachandran2007bayesian,ratliff2006maximum,taylor2011integrating,ziebart2008maximum}. To provide demonstrations on robots, a human needs to teleoperate the robot. However, this is difficult; teleoperation interfaces for robots are noisy and many have fewer DOFs than the robot itself and are therefore unintuitive to use \cite{dragan2012formalizing,javdani2015shared,khurshid2015data}. 

Additionally, IRL works best when the demonstrations are provided on a diverse set of training environments \cite{hadfield2017inverse, ratner2018simplifying}. However, it is challenging for humans to craft such an informative set of environments. Even if they were able to, it may be impractical (in terms of cost or time) to set up multiple (physical) training environments for robotics tasks.

Finally, even if the human is able to provide good demonstrations, the demonstrations themselves may not accurately describe how the human \textit{wants the robot to behave}; in other words, a person's preferences may deviate from their own actions.

Recent work found that people do not actually want their autonomous cars to drive as aggressively as they do \cite{basu2017you}. In such cases, the reward function learned by IRL will not encode the humans' desired behavior.

Imitation learning \cite{duan2017one,finn2016guided, ho2016generative, ross2011reduction} is a similar framework that uses demonstrations to train a robot. However, imitation learning directly learns a policy from the demonstrations, bypassing learning the reward function.

\prg{Preference-Based Learning.} Preference-based learning is a framework for learning reward functions that relies on querying the human for her preferences between trajectories \cite{akrour2012april,biyik2018batch,eric2008active, furnkranz2012preference, holladay2016active,jain2015learning, dorsa2017active,sugiyama2012preference,wilson2012bayesian}. The result of preference queries are more representative of the human's true reward function than demonstrations since (a) it is easier for the human to decide between two trajectories than provide a full-length demonstration, especially when she is constrained by clunky interfaces for teleoperation; and (b) the human responds according to her observation of the robot's behavior and therefore, the learned reward function will better reflect how she \textit{wants the robot to behave}. However, each preference query is much less informative than a demonstration, since they only provide information relative to one other trajectory.

Active preference-based learning methods alleviate this issue by generating the \textit{most informative} query at each step (as measured by entropy, expected volume removed, or other information theoretic metrics) \cite{ailon2012active, akrour2012april,biyik2018batch,eric2008active,dorsa2017active}. However, these methods often require solving an optimization problem over control space at each step; in robotics, where the control space is complex and high-dimensional, they tend to be inefficient.

Notably, these algorithms usually take a long time to narrow in on the rough region of space where the true reward function lies. However, once they have identified this smaller space, the algorithm is very effective; the fine-grained nature of the preference queries allow the algorithm to quickly and accurately identify the true reward function.

\prg{Using Demonstrations \& Preferences.} Recent work from \citet{ibarz2018reward} has explored combining demonstrations and preferences; specifically, they take a model-free approach to learn a (collective) reward function in the Atari domain. Our motivation (physical autonomous systems) differs from theirs, leading us to a structurally distinct method. Data from humans controlling physical robots is difficult and expensive to obtain. Therefore, model-free approaches, not as amenable to active methods, are presently impractical. In contrast, we give special attention to the issue of sample complexity. To this end, we (1) assume a simple reward structure (standard in the IRL literature) and (2) employ active methods to generate queries. Since the resulting training process is not especially time-intensive, we can efficiently learn \textit{personalized} reward functions for \textit{each} user. In our study, users trained a 7-DOF manipulator at a reaching task in $\sim$30 minutes of human time.

\section{Approach Overview and Modelling Details}

Our goal is to learn \textit{accurate} reward functions \textit{efficiently}. As previously discussed, it is difficult to get high-quality demonstrations in robotics and therefore methods that rely solely on demonstrations tend to learn inaccurate reward functions. Conversely, methods that rely exclusively on preference queries are too inefficient since the process of generating each query is computationally-intensive and only yields binary feedback on the true (continuous) reward function.

\prg{Approach Overview.} Our key insight is that: 
\begin{quote}
    \emph{Preferences queries are comparatively more accurate but less informative than demonstrations.}
\end{quote}
We leverage this insight to derive a new framework for reward learning, DemPref, that unifies IRL and preference-based learning into a single framework, amplifying the benefits of using both demonstrations and preferences while dampening their respective drawbacks.

In DemPref, we learn the reward function in two stages. In the first stage, the human provides demonstrations, which are then used to construct a probabilistic distribution over reward functions. We then use this learned distribution as a prior over reward functions in the second stage, where we actively query the human for her preference between actively-generated queries to learn the true reward function. While imperfect, this prior is more informative about the true reward function than the uniform prior which is typically used \cite{biyik2018batch, dorsa2017active}. 

In other words, we use demonstrations to quickly identify a small region of space in which the true reward function lies; we then leverage the accurate nature of preference queries to quickly zero-in on the true reward function. We specifically avoid using preference queries at the initial stages and demonstrations at the latter stages to diminish the impact of their relatively uninformative and inaccurate nature respectively.

\prg{Model.}
Consider a fully observable dynamical system describing the evolution of a robot $\Hu$, who should behave according to the preferences of a human. The continuous state is denoted by $x\in X$, the continuous control is denoted by $\uH\in \UH$ respectively. The dynamics are given by $x^{t+1} = \dyn(x^t, \uH^t).$

A trajectory, $\xi\in\Xi$, is a finite sequence of states and actions, i.e., $\xi = ((x^t, \uH^t))_{t=0}^T$, where $T$ is the horizon of the system. 

We assume that there exists a reward function, $\RH: \Xi \to \mathbb{R}$ that describes how the human wants $\Hu$ to act in this system. Our goal is to learn $\RH$. We follow prior work \cite{abbeel2004apprenticeship,dorsa2017active,ziebart2008maximum} in assuming that $\RH$ is a linear combination of features $\phi: X \times \UH \to \mathbb{R}^k$, which we assume access to. i.e., we write:$$\RH(\xi) = \textbf{w}\cdot \Phi(\xi) = \textbf{w} \cdot \sum_{t=0}^T \phi(x^t,\uH^t),$$ where we constrain the weight vector $\w$ to have $\norm{\w}_2 \leq 1$. To learn $\RH$, we simply have to learn $\textbf{w}$.

\section{DemPref: A Unified Framework \\ for Reward Learning}\label{alg_without_ic}

In the interest of clarity, we first introduce the simplest version of DemPref, which builds off the Bayesian IRL (BIRL) algorithm \cite{ramachandran2007bayesian} and a maximum volume removal algorithm for active, preference-based learning \cite{dorsa2017active}. Our contribution in this section is to assemble these two distinct approaches into one coherent framework, with consistent notation and a clear organization of ideas.

In the next section, we will discuss the rest of our framework. Specifically, we will show how to exploit the structure of DemPref to derive a modified active, preference-based algorithm which can generate better queries. We then use this modified algorithm in DemPref to learn the reward function more efficiently. The full framework is depicted in Figure \ref{schem_fig}.

\prg{Stage 1: Initializing a Reward Distribution from Demonstrations.} Let $\{\dem_{1},\ldots,\dem_{n}\}$ denote the demonstrations provided by the human. We want to derive a distribution over the weight vector $\w$ that best explains the human's demonstrations. We will do so via a Bayesian IRL approach \cite{ramachandran2007bayesian}; we thus require a probabilistic model of the human's demonstrations and a prior over the weight vector $\w$.

We assume that the demonstrations are independent (conditional on $\w$) and identically distributed; thus, we can write $$\pr(\dem_{1},\ldots,\dem_{n}\mid \w) = \prod_{i=1}^n\ \pr(\dem_{i}\mid \w).$$
As in prior work, we assume that the human is trying to maximize her reward function $\RH(\xi)$ but is unable to do so perfectly \cite{dorsa2017active,ramachandran2007bayesian,ziebart2008maximum}. We model this noisiness in human behavior by assuming that the human provides demonstrations according to a Boltzmann distribution over trajectory space (with respect to the true reward of each trajectory) \cite{dragan2013generating, ramachandran2007bayesian}:$$\pr(\dem \mid \w) \propto \exp(\beta^D \w \cdot \Phi(\dem)),$$
where $\beta^D$ is a parameter that captures the level of the human's suboptimality when providing demonstrations. 

Recall that $\w$ is constrained such that $\norm{\w}_2 \leq 1$. This, together with the fact that we have no information over $\w$ before observing any demonstrations leads to a very natural choice for a prior: uniform on the unit ball. i.e., $\w \sim Unif(\B{0}{1})$.

Armed with a probabilistic model of the human's demonstrations and a prior over $\w$, we can derive the desired distribution over $\w$ via a Bayesian update:
\begin{align*}
    \pr(\w\mid \dem_{1},\ldots,\dem_{n})
    &\propto \exp\left(\beta^D\sum_{i=1}^n\ \w\cdot\Phi(\dem_i)\right)
\end{align*}

\prg{Stage 2: Learning the Reward Function with Preference Queries.}\label{prefOld} Our goal is to generate two trajectories that will maximize the information received about the human's weight vector $\w$.
We do so via an active, maximum volume removal procedure \cite{dorsa2017active}. We first initialize a distribution over $\w$, using our learned prior from Stage 1. At each time-step, we (a) generate two trajectories and ask the human to pick between them, and then (b) use this information to update our distribution over the weight vector $\w$. These two steps are repeated until convergence. The details of these steps are presented below.

\subprg{Updating the Reward Distribution.} 
Now, assume that the human has, in response to our generated query, chosen her preferred trajectory. To incorporate the information from this query into the distribution of $\w$, we will need a model for how the human chooses her preferred trajectory. As we did in Section \ref{alg_without_ic}, we will assume that the human is noisily optimal with respect to the true reward of each trajectory. As per this model, the probability that the human selects trajectory $\xi_1$\begin{align}
    \pr(\xi_1 \mid \w) &= \frac{\exp(\beta^R \w\cdot\Phi(\xi_1)}{\sum_{i=1}^2\exp(\beta^R  \w\cdot\Phi(\xi_i))}\label{update2traj}\\
    &\approx \min(1,\ \exp(\w\cdot(\Phi(\xi_1) - \Phi(\xi_2))))\label{approx}
\end{align}
where $\beta^R$ is a parameter that captures the human's suboptimality when choosing between the trajectories. (For more detail on the approximation $\eqref{approx}$, see \cite{dorsa2017active}.) We can then update the distribution of $\w$ in a standard Bayesian fashion.

\subprg{Actively Generating Queries.} We can formulate the problem of generating maximally-informative queries as the problem of maximizing the volume removed from the (unnormalized) distribution of the weight vector $\w$: the queries are thus generated according to the following optimization problem, subject to appropriate feasibility constraints:
\begin{align}
    \max_{\xi_1, \xi_2\in \Xi} \ \min\{\mathbb{E}[1-\pr(\xi_1\mid \w)],\  \mathbb{E}[1-\pr(\xi_2\mid \w)]\}.\label{obj2traj}
\end{align}
Each term in the minimum corresponds to the volume removed from a specific preference from the human; thus, the trajectories selected by this optimization can be seen to \textit{maximize the minimum volume removed} by a query (in expectation).

\section{Iterated Correction: Generating Better Queries in DemPref} \label{alg_with_ic}

Note that the optimization problem specified by \eqref{obj2traj} is non-convex and thus, we are only guaranteed that the generated query is locally optimal in terms of the expected volume removed. In domains with a high-dimensional and complex control space, the non-convexity of the problem can lead to cases where both generated trajectories in a query are not informative of the human's true reward function, despite being locally optimal in the optimization problem. (An example of this phenomena was recently observed by \citet{biyik2018batch}.) This makes it more difficult for the human to provide useful feedback and can slow down convergence significantly.

To  alleviate this issue, we derive a simple active, preference-based learning method than can be used in DemPref. For ease of explanation, we assume for the remaining of this section that the human only provides one demonstration.

In the standard DemPref framework, the demonstration is only useful in learning a prior to generate preference queries with. Here, we additionally retain the demonstration during the second phase, and use them to ground the preference queries: at each step, the human may choose between the two generated trajectories or the demonstration she initially provided. If the human chooses one of the generated trajectories, that trajectory will take the place of the demonstration in the next set of preference queries. This ensures that there is always at least one trajectory in the query that is reasonably reflective of the human's true reward function. We can think of this method as iteratively finding better trajectories, starting from the initial demonstration; hence, we refer to this preference-based learning method as Iterated Correction (IC). We note that our method is similar in inspiration to that of \citet{jain2015learning}, where the human iteratively provides minor improvements to a trajectory proposed by the system.

We only require a few changes to the preference-learning update function \eqref{update2traj} and the the maximum volume removal objective \eqref{obj2traj} to utilize this idea in DemPref. (The rest of the algorithm follows directly from Section \ref{alg_without_ic}.)

\prg{Adapting the Update Function.} Instead of having to choose between two trajectories as in Section \ref{prefOld}, the human now has to choose between three trajectories: her demonstration/ previously chosen trajectory, which we henceforth refer to as the stored trajectory $\xi^S$, and the two new generated trajectories. We thus need to adapt our update function to account for this additional structure. (In in the interest of generality, we show how to adapt the update function in the case where the human has to choose between $n$ trajectories.)

Recall that the update function \eqref{update2traj} was derived from modeling the human as behaving noisily optimal (according to a Boltzmann distribution) with respect to the reward of each trajectory. This model can be easily extended to the case where the human is choosing between $n$ trajectories. Then, the probability that the human picks some trajectory $\xi_i$ from the set $\{\xi_0 = \xi^S, \xi_1, \ldots, \xi_{n-1}\}$ is given by:
\begin{align}
    \pr(\xi_i\mid \w) &= \frac{\exp(\beta^R \w\cdot\Phi(\xi_i)}{\sum_{j=0}^{n-1}\exp(\beta^R \w\cdot\Phi(\xi_j)}\label{update_pickbest}
\end{align}
However, note that by using this update, we would be wasting a lot of information: this update function only incorporates information about which trajectory the human prefers \textit{most}. We lose information about the human's relative preferences between the remaining trajectories, which contain more information about her true reward function. Therefore, to maximize the amount of information extracted from each query, we instead query the human for a ranking of the trajectories. 

\begin{figure}[t]
\includegraphics[width=\columnwidth]{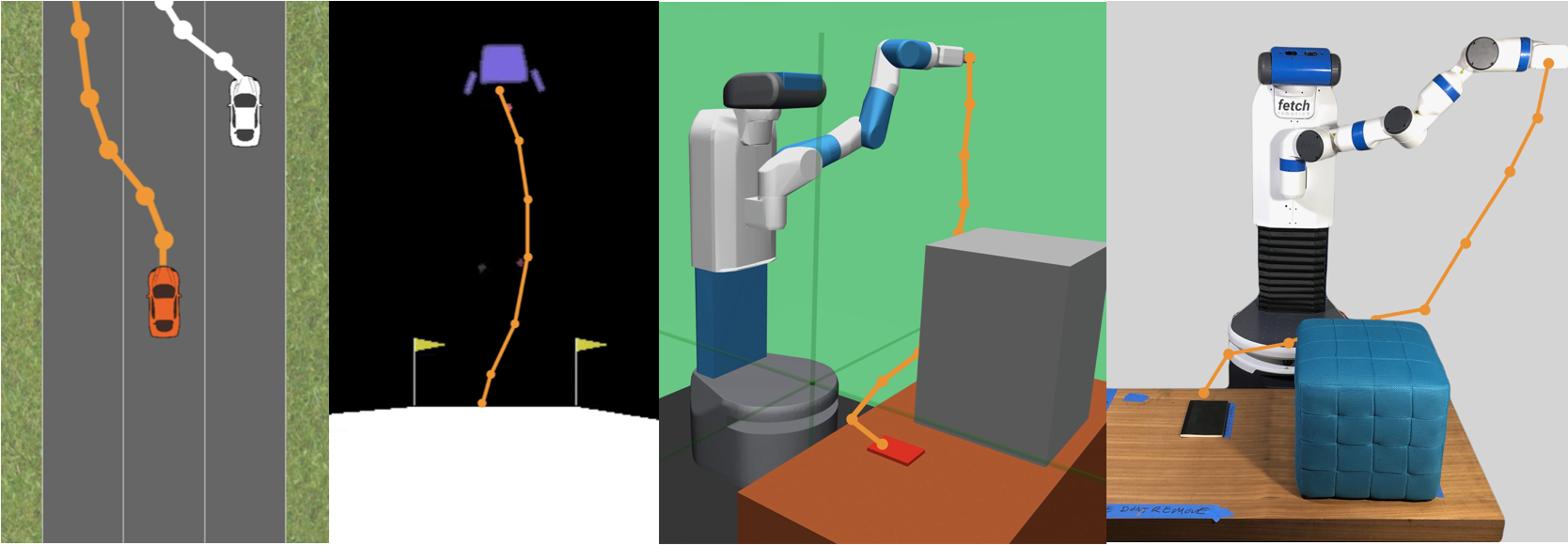}
\centering
\caption{Views from each domain, with a demonstration in orange: (a.) Driver, (b.) Lunar Lander, (c.) Fetch Reach (simulated), (d.) Fetch Reach (physical). }
\label{domains_fig}
\vspace{-0.2in}
\end{figure}

We first introduce some notation. Let $\mathbb{S}_n$ denote the algebraic group composed of length-n permutations $\sigma$; here, these permutations will map ranks to objects. e.g., if the human is presented with two trajectories, $\xi_1$ and $\xi_2$, and prefers $\xi_2$ over $\xi_1$, we can succinctly represent this information by the permutation $\sigma(1,2) = (2,1)$, i.e., $\sigma$ maps the first rank to the second object and the second rank to the first object. Thus, we denote the $i^{th}$ best trajectory in the human's ranking as $\rk{i}$. 

At each step, the human selects a ranking of all trajectories, $\rank = (\rk{1}, \ldots, \rk{n}).$ We again model the human as noisily optimal in choosing her ranking. Specifically, we assume that the human selects her ranking according to the Plackett-Luce model \cite{luce2012individual,plackett1975analysis} from discrete choice theory, which is widely applied to model noisy human behavior \cite{gormley2008exploring,guiver2009bayesian,mcfadden1973conditional}.

As per this model, the probability that the human selects some ranking $\rank = (\rk{1},\ldots,\rk{k})$ is given by:\begin{align}
    \pr(\rank\mid \w) &= \pr((\rk{1},\ldots,\rk{k})\mid \w)\nonumber\\
    =&\prod_{i=1}^k \frac{\exp(\beta^R \w\cdot\Phi(\rk{i}))}{\sum_{j=i}^k \exp(\beta^R \w\cdot\Phi(\rk{j}))} \label{update_rank}
\end{align}

\prg{Adapting the Objective Function.} As above, we will show how to adapt the objective function to generate a query of $n$ trajectories (one of which is the stored trajectory, $\xi^S$). To maximize the information gained, we generate these new $n-1$ trajectories, $\xi_1, \ldots, \xi_{n-1}$ to maximize the expected volume \textit{jointly removed by the stored trajectory and the newly generated trajectories}. We adapt \eqref{obj2traj} to do so:
\begin{align}
\max_{\xi_1, \ldots, \xi_{n-1}\in \Xi}\ \min_{\sigma\in\mathbb{S}_n} \ \{\mathbb{E}[1-\pr(\sigma\mid \w)]\}.\label{obj_rank}
\end{align}
We then query the human with the stored trajectory and all the newly generated trajectories: $\{\xi^S, \xi_1, \ldots, \xi_{n-1}\}.$ We then update the distribution of $\w$ with this information and set the new stored trajectory to be human's most preferred trajectory $\rk{1}$, before generating a new set of queries.

\prg{Iterated Correction with Multiple Demonstrations.} It is straightforward to use this method even when there are multiple demonstrations provided by the human. We simply store all the demonstrations in a buffer and at each time step, we randomly draw a demonstration to use in the next query; at the end of each iteration, the human's most preferred trajectory replaces the selected demonstration in the buffer.

\section{Simulation Experiments}\label{domains}

In this section, we run three different numerical experiments in simulation, to evaluate the effectiveness of DemPref and our related contributions. We first discuss some experimental details that are common across all experiments.

\prg{Domains.} In each experiment, we use a subset of the following domains, shown in Figure \ref{domains_fig}:

\textit{Driver}: We use a 2D driving simulator \cite{dorsa2017active}, where the agent has to safely drive down a highway. The features of the reward function in this domain correspond to the distance of the agent from the center of the lane, its speed, heading angle, and distance to other vehicles (white in Figure \ref{domains_fig}.a.).

\textit{Lunar Lander}: We use the continuous LunarLander environment from OpenAI Gym \cite{brockman2016openai}, where the lander has to safely reach the landing pad. The features in this domain correspond to the lander's average distance from the landing pad, its angle, its velocity, and its final distance to the landing pad. 

\textit{Fetch Reach}: We use a modification of the Fetch Reach environment from OpenAI Gym \cite{brockman2016openai} (built on top of MuJoCo \cite{todorov2012mujoco}), where the robot has to reach a goal with its arm, while keeping its arm as low-down as possible. The features in this domain correspond to the robot gripper's average distance to the goal, its average height from the table, and its average distance to a box obstacle in the domain.

\prg{Dependent Measure for Simulation Experiments.} For each of the above domains, we choose a weight vector $\w_{true}$, which we take to be the human's true reward function. We measure the convergence of our algorithm by computing the expected cosine similarity between $\w_{true}$ and the learned distribution over weight vectors $p(\w)$:
$$m = \mathbb{E}_{\w\sim p(\w)}\left[\frac{\w\cdot\w_{true}}{||\w||_2 ||\w_{true}||_2}\right]$$
Note that $m = 1$ implies that learned distribution $p(\w)$ has perfectly converged to the true weight vector $\w_{true}.$

\prg{Generating Demonstrations in Simulation.} Where required, we generate demonstrations in simulation via model predictive control (MPC). We do so by solving the following optimization problem: $\max_{u_1, \ldots, u_T} \w_{true} \cdot\Phi(\xi), \label{mpc}$, where $\w_{true}$ corresponds to the true weight vector.

\begin{figure*}[t]
\includegraphics[width=\textwidth]{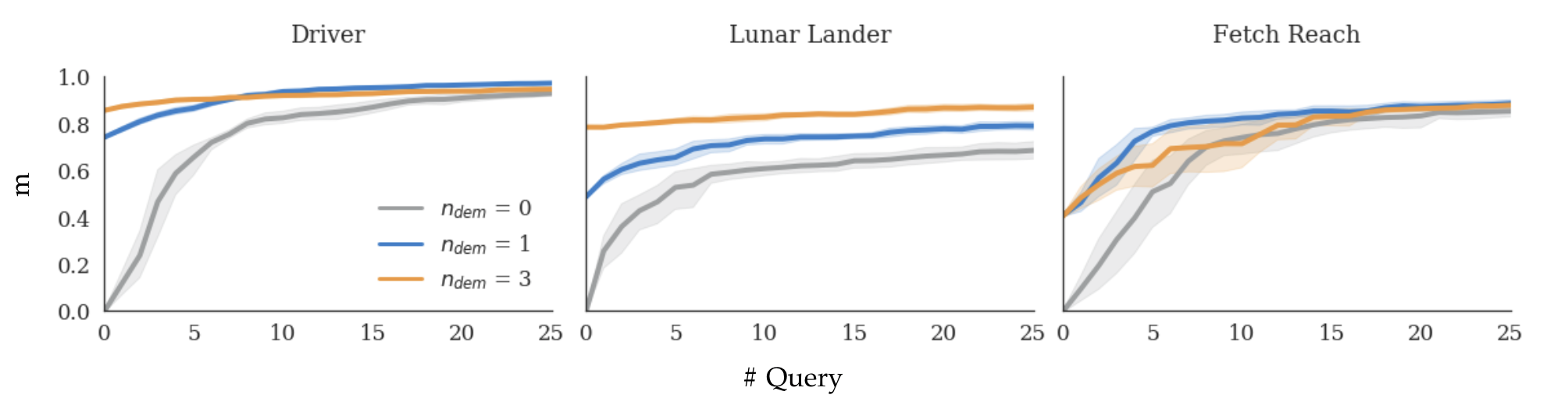}
\centering
\vspace{-0.3in}
\caption{The results of our first experiment, investigating whether initializing with demonstrations improves convergence of the algorithm, on all three domains. On the Driver, Lunar Lander, and Fetch Reach environments, initializing with one demonstration improved the rate of convergence significantly.}
\label{exp1}
\vspace{-0.2in}
\end{figure*}

\subsection{The Impact of Initializing with Demonstrations.} 
We investigate whether our standard DemPref algorithm (without IC) is more effective at learning the true weight vector than the standard active preference-based learning algorithm of \citet{dorsa2017active}, i.e., we invetigate whether initializing with demonstrations helps convergence.

\prg{Hypotheses.}

\textit{\textbf{H1.} The preference-based learning algorithm converges to the true reward faster if initialized with demonstrations.}

\textit{\textbf{H2.} The convergence of the algorithm improves with the number of demonstrations used to initialize the algorithm.
}

\prg{Manipulated Variables.} We vary the number of demonstrations, $n_{dem}$, used to initialize the algorithm across the values $\{0, 1, 3\}.$ $n_{dem} = 0$ corresponds to the standard active preference-based learning algorithm.

\prg{Experimental Setup.} We run this experiment on all three domains. We run each condition 8 times to generate confidence intervals for our results. (We do the same for all simulation experiments.) We use the DemPref algorithm with $n_{dem}$ demonstrations and 25 queries. We do not use IC to avoid confounding the results. Additional experimental details for this and all experiments can be found in the supplementary material and in the provided code.

\prg{Results and Analysis.} The results of the experiment are presented in Figure \ref{exp1}. On all three environments, using DemPref and initializing with demonstrations improves the convergence rate of the preference-based algorithm significantly; to match the $m$ value attained by the algorithm initialized with only one demonstration in 10 preference queries, it takes the standard preference-based algorithm ~30 queries on Driver, ~35 queries on Lander, and ~20 queries on Fetch Reach. These results provide strong evidence in favor of H1.

The results regarding H2 are more complicated. Initializing with three instead of one demonstrations only improves convergence significantly on the Driver and Lunar Lander domains. (The improvement on Driver is only at the early stages of the algorithm, when fewer than 10 preferences are used.) However, on the Fetch Reach domain, initializing with three instead of one demonstration hurt the performance of the algorithm. (Although, we do note that the results from using three demonstrations are still an improvement over the results from not initializing with demonstrations). This is unsurprising. It is much harder to provide demonstrations on the Fetch Reach environment than on the Driver or Lunar Lander environments, and therefore the demonstrations are of lower quality. Using more demonstrations when they are of lower quality leads to the prior being more concentrated further away from the true reward function, and can cause the preference-based learning algorithm to slow down.

In practice, we find that using a single demonstration to initialize the algorithm leads to reliable improvements in convergence, regardless of the complexity of the domain. (Indeed, in the rest of the simulation experiments and the user study, we only initialize the algorithm with one demonstration.)

\subsection{The Impact of Using Rankings instead of Preferences.}\label{exp2_sec} 

We investigate whether our ranking update \eqref{update_rank} improves the convergence of the preference-based learning algorithm.

\prg{Hypotheses.}

\textit{\textbf{H3.} The preference-based learning algorithm converges to the true reward function faster when using the ranking update \eqref{update_rank} instead of the preference update \eqref{update_pickbest}.}

\prg{Manipulated Variables.} We vary the update function used, between the preference update (Pref) and the ranking update (Rank). Additionally, we vary the number of options presented to the user, $n_{opt}$, at each step across the values $\{3, 5\}$.

\prg{Experimental Setup.} We run this experiment on the Driver and Lunar Lander domains.We use the standard preference-based learning algorithm with 25 preference queries. We do not initialize with demonstrations or use Iterated Correction to avoid confounding the results.

\begin{figure}[t]
\includegraphics[width=\columnwidth]{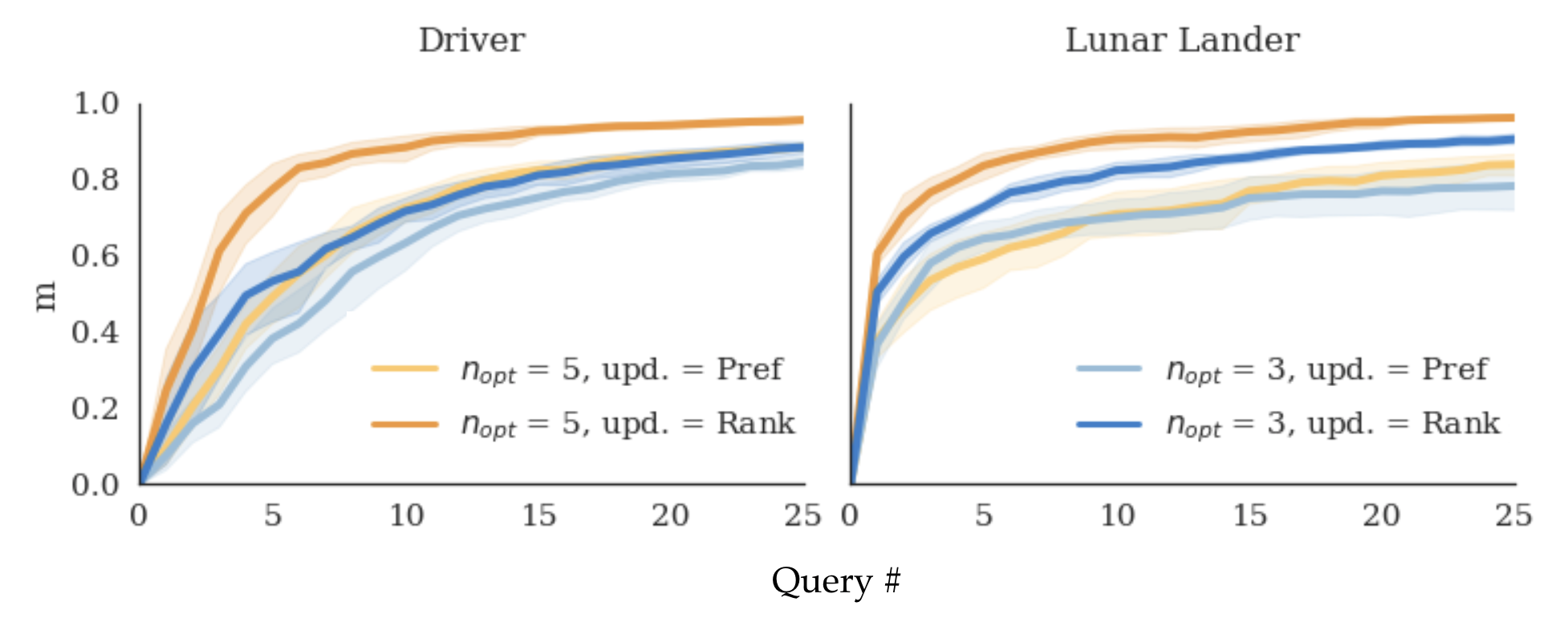}
\centering
\caption{The results of our second experiment, investigating whether our ranking update improves convergence of the algorithm over the standard preference update. The algorithm with the ranking update significantly outperformed the algorithm with the preference update for each value of $n_{opt}$, on both domains.}
\label{exp2}
\vspace{-0.2in}
\end{figure}

\prg{Results and Analysis.} The results of this experiment are presented in Figure \ref{exp2}. For both values of $n_{opt}$, the algorithm with the ranking update significantly outperforms its counterpart with the preference update. Additionally, we note that the ranking update-algorithm that presents the user with 3 options does \textit{as well as} the preference update-algorithm that presents the user with 5 options. These results strongly support H3. The ranking update allows us to maximize the amount of information extracted from the human from each query, which reduces the number of queries (i.e., computation) needed.

\subsection{The Impact of Iterated Correction.} Finally, we investigate whether using DemPref with IC instead of with the standard active preference-based learning method, improves convergence. Additionally, we investigate how the effectiveness of the method relates to the quality of the initial demonstration provided.

\prg{Hypotheses.}

\textit{\textbf{H4.} The algorithm converges to the true reward function faster when using the Iterated Correction method.}

\textit{\textbf{H5.} The improvement in convergence resulting from the Iterated Correction method is greater when the initial demonstration is not of high quality.
}

\prg{Manipulated Variables.} We vary (a) whether the IC method is used and (b) the quality of the demonstration used between being of low- and high-quality.

\prg{Experimental Setup.} We run this experiment on the Driver and Lunar Lander domains.
We use our algorithm with 1 demonstration and 25 preference queries. We generate the low- and high-quality demonstrations by generating 100 demonstrations and picking the worst and best (as measured by the convergence metric) demonstration respectively.

\begin{figure}[t]
\includegraphics[width=\columnwidth]{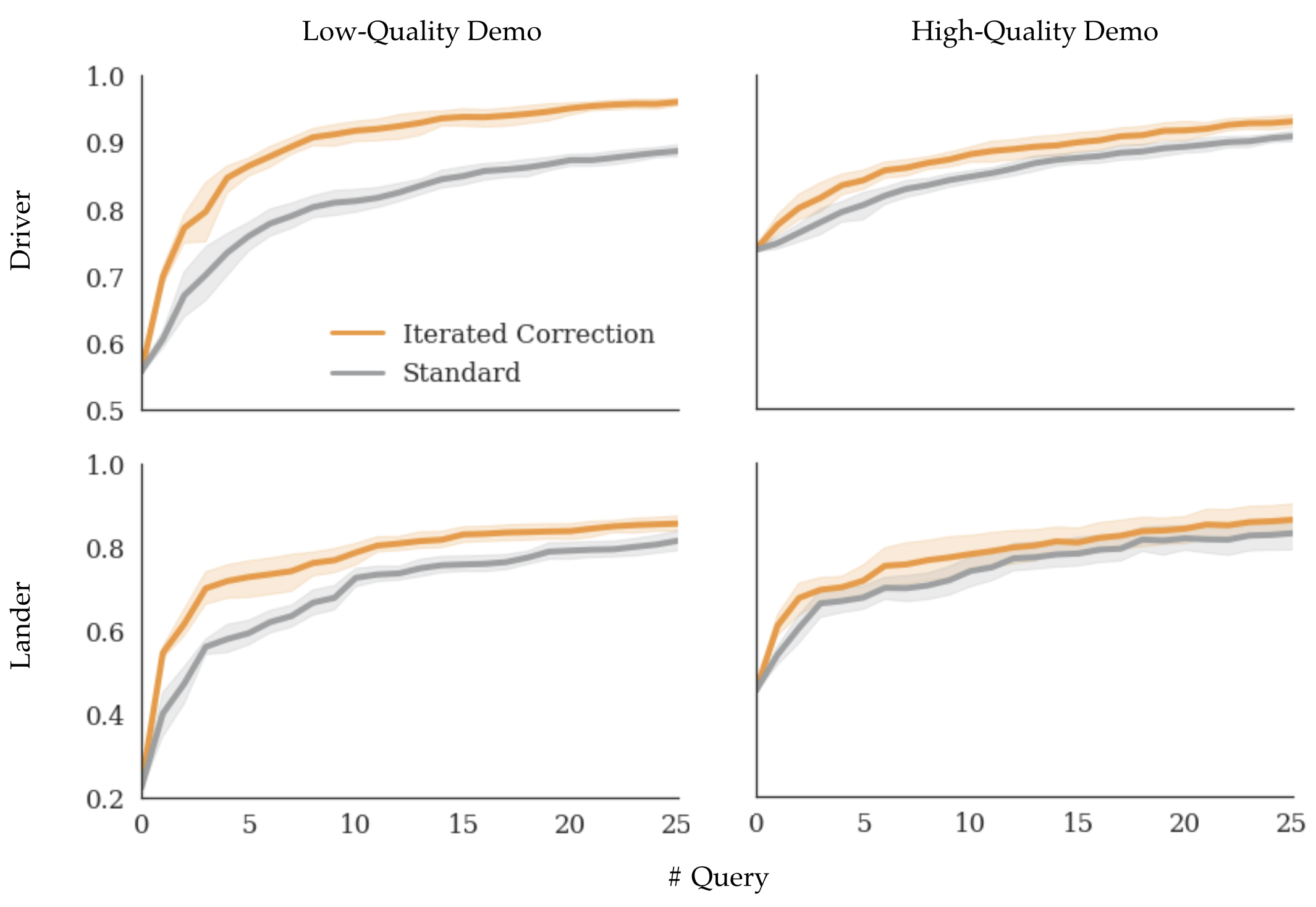}
\centering
\caption{The results of our third experiment, investigating whether using our Iterated Correction method instead of a standard active preference-based learning method improves the convergence of the full algorithm. In 3 of the 4 settings, we find that Iterated Correction improves the convergence of the algorithm; we additionally find that the relative benefit of using Iterated Correction is greater when the initial demonstration is of lower quality.}
\label{exp3}
\vspace{-0.2in}
\end{figure}

\prg{Results and Analysis.} The results of this experiment are presented in Figure \ref{exp3}. IC improves the convergence of DemPref significantly in both cases on the Driver domain, and in the case with the lower-quality demonstration on the Lunar Lander domain. Even on the remaining case in the Lunar Lander environment, IC still improves the convergence of the algorithm, although the result is not statistically significant. These results provide some evidence in favor of H4.

We additionally note that in both domains, the improvement in convergence resulting from using Iterated Correction is greater when the quality of the initial demonstration is low. When using a low-quality demonstration, to match the $m$ value attained by the algorithm with Iterated Correction in 10 preference queries, it takes the algorithm without Iterated Correction $\gg$25 queries on Driver and $\sim$20 queries on Lander. When using a high-quality demonstration, it takes $\sim$20 queries on Driver and $\sim$15 queries on Lander. These results provide strong evidence in favor of H5.

This latter result may seem surprising but is actually far from being so. Recall that the preference-based learning algorithm uses a maximum volume removal approach to generating queries, i.e., it attempts to generate as diverse as possible a set of queries. If the initial demonstration is already of high-quality (i.e., it is already very reflective of the true reward function), higher-quality trajectories are likely to be similar to the demonstration and unlikely to be generated by the algorithm. However, this is not a problem if the initial demonstration is of low(er) quality; the algorithm can more easily generate better trajectories since they are less likely to be similar to the demonstration.

\section{User Study}\label{user_study}
\begin{figure*}[t]
\includegraphics[width=\textwidth]{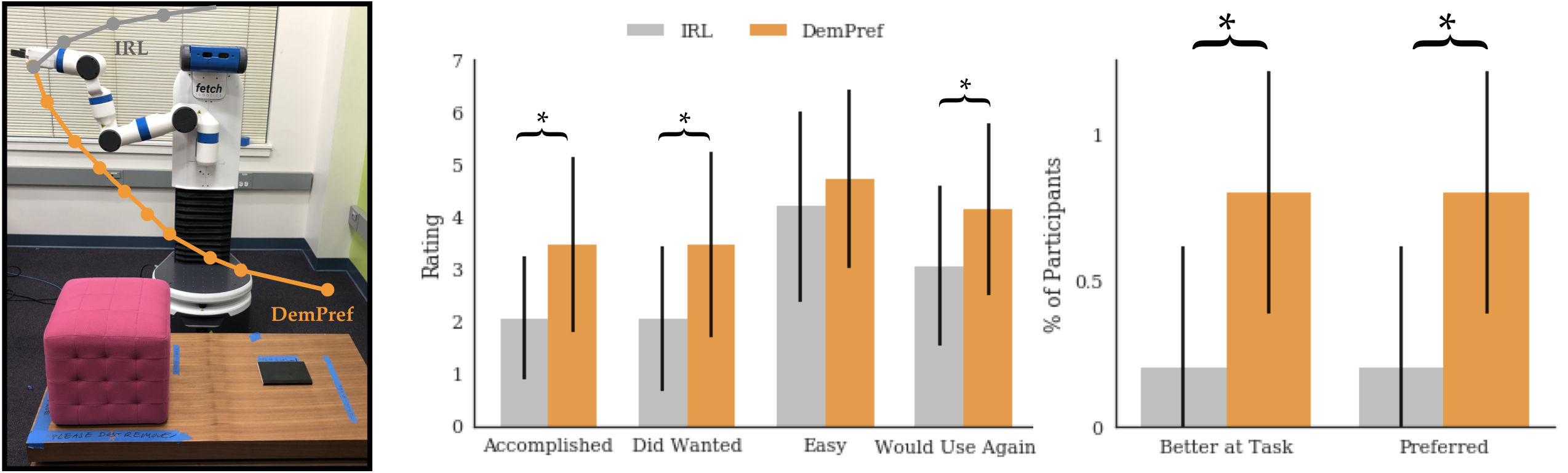}
\centering
\caption{(Left) Our testing domain, with two trajectories generated according to the reward functions learnt by IRL and DemPref from a specific user in our study. (Right) The results of our usability study -- the error bars correspond to standard deviation and significant results are marked with an asterisk. We find that users rated the robot trained with DemPref as significantly better at accomplishing the task and preferred to use our method for training the robot significantly more than they did IRL. However, we did not find evidence to suggest that users found our method easier to use than standard IRL.}
\label{human_exp_fig}
\vspace{-0.2in}
\end{figure*}

In this section, we empirically evaluate the effectiveness of DemPref at learning human reward functions, against that of a standard IRL method by conducting a within-subjects study. 

\prg{Experimental Setup.} We use the Fetch Reach domain described in Section \ref{domains}, except with a physical Fetch robot. Participants trained the robot using two different systems (described below). They were told that their goal was to get the robot's end-effector as close as possible to the goal, while (1) avoiding collisions with the block obstacle and (2) keeping the robot's end-effector low to the ground (so as to avoid, for example, knocking over objects around it). (Our experimental domain was inspired by that of \citet{bajcsy2017learning}.) Demonstrations, where required, were provided via teleoperation (using end-effector control) on a keyboard interface; each user was given some time to familiarize themselves with the teleoperation system before beginning the experiment. The training domain for this task is shown in Figures \ref{front_fig} and \ref{domains_fig}.d. 

After training, the robot was trained in simulation using Proximal Policy Optimization (PPO) with the reward function learned from each system \cite{baselines, schulman2017proximal}. We rolled out three trajectories for each system on the physical Fetch, and the users were asked to evaluate the robot's behavior.

To ensure that the robot wasn't simply overfitting to the training domain, we used different variants of the domain for training and testing the robot. We used two different test domains (and counter-balanced across them) to increase the robustness of our results against the specific testing domain. Figure \ref{human_exp_fig} (left) illustrates one of our testing domains.

\prg{Manipulated Variables.} Participants used two different systems to train the robot. (1) IRL: Bayesian IRL  with 5 demonstrations. (2) DemPref: our DemPref framework (without IC) with 1 demonstration and 15 preference queries. (The number of demonstrations and preferences used in each system were chosen such that a simulated agent achieves similar convergence to the true reward on both systems.) We counter-balanced across which system was used first to minimize the impact of familiarity bias with our teleoperation system.

\prg{Dependent Measures.} For each condition, the users were shown three trajectories generated according to the reward function learned from that system, on the physical Fetch. After observing each set of trajectories, the users were asked to rate the following statements on a 7-point Likert scale:\begin{enumerate}
    \item The robot accomplished the task well. (Accomplished)
    \item The robot did what I wanted. (Did Wanted)
    \item It was easy to train the robot with this system. (Easy)
    \item I would want to use this system to train a robot in the future. (Would Use Again)
\end{enumerate}
Finally, they were asked two comparison questions:
\begin{enumerate}
    \item Which robot accomplished the task better? (Better at Task)
    \item Which system would you prefer to use if you had to train a robot to accomplish a similar task? (Preferred)
\end{enumerate}
They were finally asked for general comments.

\prg{Hypothesis.}

\textit{\textbf{H1.} The DemPref-robot will be more successful at the task (as evaluated by the users) than the IRL-robot.}

\textit{\textbf{H2.} Participants will prefer to use the DemPref system as opposed to the IRL system.}

\prg{Subject Allocation.} We recruited \textbf{15} participants (11 male, 4 female), six of whom had prior experience in robotics but none of whom had any prior exposure to our system.

\prg{Videos.} In the supplementary material, we provide videos of our system in action. The accompanied PDF document contains a description and a brief discussion of each video.

\prg{Results, Analysis, and Discussion.} We present our results in Figure \ref{human_exp_fig} (right). To test our hypotheses, we ran a Wilcoxon paired signed-rank test on the comparison questions. When asked which robot accomplished the task better, users preferred the DemPref system by a significant margin ($p = 0.0201$); similarly, when asked which system they would prefer to use in the future if they had to train the robot, users preferred the DemPref system by a significant margin ($p = 0.0201$). This provides strong evidence in favor of both \textbf{H1} and \textbf{H2}. 

As expected, many users struggled to teleoperate the robot. Several users made explicit note of this fact in their comments: ``I had a hard time controlling the robot", ``I found the [IRL system] difficult as someone who [is not] kinetically gifted!", ``Would be nice if the controller for the [robot] was easier to use." Given that the IRL-robot was only trained on these demonstrations, it is perhaps unsurprising that the DemPref-robot outperforms the IRL-robot on the task.

We were however surprised by the extent to which the IRL-robot fared poorly: in many cases, the IRL-robot did not even attempt to reach for the goal. Upon further investigation, we discovered that IRL was prone to, in essence, ``overfitting" to the training domain. In several cases, IRL had overweighted the users' preference for collision avoidance. This proved to be an issue in one of our test domains where the obstacle is closer to the robot than in the training domain. Here, the robot does not even try to reach for the goal since the loss in value (as measured by the learned reward function) from going near the obstacle is greater than the gain in value from reaching for the goal. Figure \ref{human_exp_fig} (left) shows this training domain and illustrates, for a specific user, a trajectory generated according to reward function learned by each of IRL and DemPref. Videos of these trajectories are provided in the supplementary material as well. 

While we expect that IRL would overcome these issues with more careful feature engineering and increased diversity of the training domains, it is worth noting that our method was affected much less by these issues. These results suggest that preference-based learning methods may be more robust to poor feature engineering and a lack of training diversity than IRL; however, a rigorous evaluation of these claims is beyond the scope of this paper and we leave it for future work. 

It is interesting that despite the challenges that users faced with teleoperating the robot, they did not rate the DemPref system as being ``easier" to use than the IRL system ($p = 0.297$). Several users specifically referred to the time it took to generate each query ($\sim$45 seconds) as negatively impacting their experience with the DemPref system: ``I wish it was faster to generate the preference [queries]", ``The [DemPref system] will be even better if time cost is less." Additionally, one user expressed difficulty in evaluating the preference queries themselves, commenting that ``It was tricky to understand/infer what the preferences were [asking]. Would be nice to communicate that somehow to the user (e.g. which [trajectory] avoids collision better)!" These are valid critiques of the framework, and indeed, we are currently investigating approaches to (a) generate preferences more quickly via distributed optimization \cite{boyd2011distributed} and to (b) generate more interpretable preferences as one direction for future work.

\section*{Acknowledgments}
Toyota Research Institute (``TRI")  provided funds to assist the authors with their research but this article solely reflects the opinions and conclusions of its authors and not TRI or any other Toyota entity. This project was also supported by FLI grant RFP2-000.

\bibliographystyle{plainnat}
\bibliography{references}

\begin{thebibliography}{46}
\providecommand{\natexlab}[1]{#1}
\providecommand{\url}[1]{\texttt{#1}}
\expandafter\ifx\csname urlstyle\endcsname\relax
  \providecommand{\doi}[1]{doi: #1}\else
  \providecommand{\doi}{doi: \begingroup \urlstyle{rm}\Url}\fi

\bibitem[Abbeel and Ng(2004)]{abbeel2004apprenticeship}
Pieter Abbeel and Andrew~Y Ng.
\newblock Apprenticeship learning via inverse reinforcement learning.
\newblock In \emph{Proceedings of the twenty-first international conference on
  Machine learning}, page~1. ACM, 2004.

\bibitem[Ailon(2012)]{ailon2012active}
Nir Ailon.
\newblock An active learning algorithm for ranking from pairwise preferences
  with an almost optimal query complexity.
\newblock \emph{Journal of Machine Learning Research}, 13\penalty0
  (Jan):\penalty0 137--164, 2012.

\bibitem[Akrour et~al.(2012)Akrour, Schoenauer, and Sebag]{akrour2012april}
Riad Akrour, Marc Schoenauer, and Mich{\`e}le Sebag.
\newblock April: Active preference learning-based reinforcement learning.
\newblock In \emph{Joint European Conference on Machine Learning and Knowledge
  Discovery in Databases}, pages 116--131. Springer, 2012.

\bibitem[Argall et~al.(2009)Argall, Chernova, Veloso, and
  Browning]{argall2009survey}
Brenna~D Argall, Sonia Chernova, Manuela Veloso, and Brett Browning.
\newblock A survey of robot learning from demonstration.
\newblock \emph{Robotics and autonomous systems}, 57\penalty0 (5):\penalty0
  469--483, 2009.

\bibitem[Bajcsy et~al.(2017)Bajcsy, Losey, O’Malley, and
  Dragan]{bajcsy2017learning}
Andrea Bajcsy, Dylan~P Losey, Marcia~K O’Malley, and Anca~D Dragan.
\newblock Learning robot objectives from physical human interaction.
\newblock In \emph{Conference on Robot Learning}, pages 217--226, 2017.

\bibitem[Basu et~al.(2017)Basu, Yang, Hungerman, Singhal, and
  Dragan]{basu2017you}
Chandrayee Basu, Qian Yang, David Hungerman, Mukesh Singhal, and Anca~D Dragan.
\newblock Do you want your autonomous car to drive like you?
\newblock In \emph{Proceedings of the 2017 ACM/IEEE International Conference on
  Human-Robot Interaction}, pages 417--425. ACM, 2017.

\bibitem[B{\i}y{\i}k and Sadigh(2018)]{biyik2018batch}
Erdem B{\i}y{\i}k and Dorsa Sadigh.
\newblock Batch active preference-based learning of reward functions.
\newblock \emph{arXiv preprint arXiv:1810.04303}, 2018.

\bibitem[Boyd et~al.(2011)Boyd, Parikh, Chu, Peleato, Eckstein,
  et~al.]{boyd2011distributed}
Stephen Boyd, Neal Parikh, Eric Chu, Borja Peleato, Jonathan Eckstein, et~al.
\newblock Distributed optimization and statistical learning via the alternating
  direction method of multipliers.
\newblock \emph{Foundations and Trends{\textregistered} in Machine learning},
  3\penalty0 (1):\penalty0 1--122, 2011.

\bibitem[Brockman et~al.(2016)Brockman, Cheung, Pettersson, Schneider,
  Schulman, Tang, and Zaremba]{brockman2016openai}
Greg Brockman, Vicki Cheung, Ludwig Pettersson, Jonas Schneider, John Schulman,
  Jie Tang, and Wojciech Zaremba.
\newblock Openai gym.
\newblock \emph{arXiv preprint arXiv:1606.01540}, 2016.

\bibitem[Brown et~al.(2019)Brown, Cui, and Niekum]{brown2019risk}
Daniel~S Brown, Yuchen Cui, and Scott Niekum.
\newblock Risk-aware active inverse reinforcement learning.
\newblock \emph{arXiv preprint arXiv:1901.02161}, 2019.

\bibitem[Chopra(2018)]{chopra_2018}
Joy Chopra.
\newblock Make the table/big block in fetch environments fixed., 2018.
\newblock URL \url{https://github.com/openai/gym/issues/920}.

\bibitem[Cui and Niekum(2018)]{cui2018active}
Yuchen Cui and Scott Niekum.
\newblock Active reward learning from critiques.
\newblock In \emph{2018 IEEE International Conference on Robotics and
  Automation (ICRA)}, pages 6907--6914. IEEE, 2018.

\bibitem[Dhariwal et~al.(2017)Dhariwal, Hesse, Klimov, Nichol, Plappert,
  Radford, Schulman, Sidor, Wu, and Zhokhov]{baselines}
Prafulla Dhariwal, Christopher Hesse, Oleg Klimov, Alex Nichol, Matthias
  Plappert, Alec Radford, John Schulman, Szymon Sidor, Yuhuai Wu, and Peter
  Zhokhov.
\newblock Openai baselines.
\newblock \url{https://github.com/openai/baselines}, 2017.

\bibitem[Dragan and Srinivasa(2013)]{dragan2013generating}
Anca Dragan and Siddhartha Srinivasa.
\newblock Generating legible motion.
\newblock 2013.

\bibitem[Dragan and Srinivasa(2012)]{dragan2012formalizing}
Anca~D Dragan and Siddhartha~S Srinivasa.
\newblock \emph{Formalizing assistive teleoperation}.
\newblock MIT Press, July, 2012.

\bibitem[Duan et~al.(2017)Duan, Andrychowicz, Stadie, Ho, Schneider, Sutskever,
  Abbeel, and Zaremba]{duan2017one}
Yan Duan, Marcin Andrychowicz, Bradly Stadie, OpenAI~Jonathan Ho, Jonas
  Schneider, Ilya Sutskever, Pieter Abbeel, and Wojciech Zaremba.
\newblock One-shot imitation learning.
\newblock In \emph{Advances in neural information processing systems}, pages
  1087--1098, 2017.

\bibitem[Eric et~al.(2008)Eric, Freitas, and Ghosh]{eric2008active}
Brochu Eric, Nando~D Freitas, and Abhijeet Ghosh.
\newblock Active preference learning with discrete choice data.
\newblock In \emph{Advances in neural information processing systems}, pages
  409--416, 2008.

\bibitem[Finn et~al.(2016)Finn, Levine, and Abbeel]{finn2016guided}
Chelsea Finn, Sergey Levine, and Pieter Abbeel.
\newblock Guided cost learning: Deep inverse optimal control via policy
  optimization.
\newblock In \emph{International Conference on Machine Learning}, pages 49--58,
  2016.

\bibitem[F{\"u}rnkranz et~al.(2012)F{\"u}rnkranz, H{\"u}llermeier, Cheng, and
  Park]{furnkranz2012preference}
Johannes F{\"u}rnkranz, Eyke H{\"u}llermeier, Weiwei Cheng, and Sang-Hyeun
  Park.
\newblock Preference-based reinforcement learning: a formal framework and a
  policy iteration algorithm.
\newblock \emph{Machine learning}, 89\penalty0 (1-2):\penalty0 123--156, 2012.

\bibitem[Gormley and Murphy(2008)]{gormley2008exploring}
Isobel~Claire Gormley and Thomas~Brendan Murphy.
\newblock Exploring voting blocs within the irish electorate: A mixture
  modeling approach.
\newblock \emph{Journal of the American Statistical Association}, 103\penalty0
  (483):\penalty0 1014--1027, 2008.

\bibitem[Guiver and Snelson(2009)]{guiver2009bayesian}
John Guiver and Edward Snelson.
\newblock Bayesian inference for plackett-luce ranking models.
\newblock In \emph{proceedings of the 26th annual international conference on
  machine learning}, pages 377--384. ACM, 2009.

\bibitem[Hadfield-Menell et~al.(2017)Hadfield-Menell, Milli, Abbeel, Russell,
  and Dragan]{hadfield2017inverse}
Dylan Hadfield-Menell, Smitha Milli, Pieter Abbeel, Stuart~J Russell, and Anca
  Dragan.
\newblock Inverse reward design.
\newblock In \emph{Advances in Neural Information Processing Systems}, pages
  6765--6774, 2017.

\bibitem[Ho and Ermon(2016)]{ho2016generative}
Jonathan Ho and Stefano Ermon.
\newblock Generative adversarial imitation learning.
\newblock In \emph{Advances in Neural Information Processing Systems}, pages
  4565--4573, 2016.

\bibitem[Holladay et~al.(2016)Holladay, Javdani, Dragan, and
  Srinivasa]{holladay2016active}
Rachel Holladay, Shervin Javdani, Anca Dragan, and Siddhartha Srinivasa.
\newblock Active comparison based learning incorporating user uncertainty and
  noise.
\newblock In \emph{RSS Workshop on Model Learning for Human-Robot
  Communication}, 2016.

\bibitem[Ibarz et~al.(2018)Ibarz, Leike, Pohlen, Irving, Legg, and
  Amodei]{ibarz2018reward}
Borja Ibarz, Jan Leike, Tobias Pohlen, Geoffrey Irving, Shane Legg, and Dario
  Amodei.
\newblock Reward learning from human preferences and demonstrations in atari.
\newblock In \emph{Advances in Neural Information Processing Systems}, pages
  8022--8034, 2018.

\bibitem[Jain et~al.(2015)Jain, Sharma, Joachims, and Saxena]{jain2015learning}
Ashesh Jain, Shikhar Sharma, Thorsten Joachims, and Ashutosh Saxena.
\newblock Learning preferences for manipulation tasks from online coactive
  feedback.
\newblock \emph{The International Journal of Robotics Research}, 34\penalty0
  (10):\penalty0 1296--1313, 2015.

\bibitem[Javdani et~al.(2015)Javdani, Srinivasa, and
  Bagnell]{javdani2015shared}
Shervin Javdani, Siddhartha~S Srinivasa, and J~Andrew Bagnell.
\newblock Shared autonomy via hindsight optimization.
\newblock \emph{arXiv preprint arXiv:1503.07619}, 2015.

\bibitem[Kelcey(2017)]{kelcey_2017}
Mat Kelcey, 2017.
\newblock URL \url{https://twitter.com/mat_kelcey/status/886101319559335936}.

\bibitem[Khurshid and Kuchenbecker(2015)]{khurshid2015data}
Rebecca~P Khurshid and Katherine~J Kuchenbecker.
\newblock Data-driven motion mappings improve transparency in teleoperation.
\newblock \emph{Presence: Teleoperators and Virtual Environments}, 24\penalty0
  (2):\penalty0 132--154, 2015.

\bibitem[Levine and Koltun(2012)]{levine2012continuous}
Sergey Levine and Vladlen Koltun.
\newblock Continuous inverse optimal control with locally optimal examples.
\newblock \emph{arXiv preprint arXiv:1206.4617}, 2012.

\bibitem[Luce(2012)]{luce2012individual}
R~Duncan Luce.
\newblock \emph{Individual choice behavior: A theoretical analysis}.
\newblock Courier Corporation, 2012.

\bibitem[Malik et~al.(2018)Malik, Palaniappan, Fisac, Hadfield-Menell, Russell,
  and Dragan]{malik2018efficient}
Dhruv Malik, Malayandi Palaniappan, Jaime~F Fisac, Dylan Hadfield-Menell,
  Stuart Russell, and Anca~D Dragan.
\newblock An efficient, generalized bellman update for cooperative inverse
  reinforcement learning.
\newblock \emph{arXiv preprint arXiv:1806.03820}, 2018.

\bibitem[McFadden et~al.(1973)]{mcfadden1973conditional}
Daniel McFadden et~al.
\newblock Conditional logit analysis of qualitative choice behavior.
\newblock 1973.

\bibitem[Plackett(1975)]{plackett1975analysis}
Robin~L Plackett.
\newblock The analysis of permutations.
\newblock \emph{Applied Statistics}, pages 193--202, 1975.

\bibitem[Ramachandran and Amir(2007)]{ramachandran2007bayesian}
Deepak Ramachandran and Eyal Amir.
\newblock Bayesian inverse reinforcement learning.
\newblock \emph{Urbana}, 51\penalty0 (61801):\penalty0 1--4, 2007.

\bibitem[Ratliff et~al.(2006)Ratliff, Bagnell, and
  Zinkevich]{ratliff2006maximum}
Nathan~D Ratliff, J~Andrew Bagnell, and Martin~A Zinkevich.
\newblock Maximum margin planning.
\newblock In \emph{Proceedings of the 23rd international conference on Machine
  learning}, pages 729--736. ACM, 2006.

\bibitem[Ratner et~al.(2018)Ratner, Hadfield-Menell, and
  Dragan]{ratner2018simplifying}
Ellis Ratner, Dylan Hadfield-Menell, and Anca~D Dragan.
\newblock Simplifying reward design through divide-and-conquer.
\newblock \emph{arXiv preprint arXiv:1806.02501}, 2018.

\bibitem[Ross et~al.(2011)Ross, Gordon, and Bagnell]{ross2011reduction}
St{\'e}phane Ross, Geoffrey Gordon, and Drew Bagnell.
\newblock A reduction of imitation learning and structured prediction to
  no-regret online learning.
\newblock In \emph{Proceedings of the fourteenth international conference on
  artificial intelligence and statistics}, pages 627--635, 2011.

\bibitem[Sadigh et~al.()Sadigh, Dragan, Sastry, and Seshia]{dorsa2017active}
Dorsa Sadigh, Anca~D Dragan, Shankar Sastry, and Sanjit~A Seshia.
\newblock Active preference-based learning of reward functions.

\bibitem[Schulman et~al.(2017)Schulman, Wolski, Dhariwal, Radford, and
  Klimov]{schulman2017proximal}
John Schulman, Filip Wolski, Prafulla Dhariwal, Alec Radford, and Oleg Klimov.
\newblock Proximal policy optimization algorithms.
\newblock \emph{arXiv preprint arXiv:1707.06347}, 2017.

\bibitem[Sugiyama et~al.(2012)Sugiyama, Meguro, and
  Minami]{sugiyama2012preference}
Hiroaki Sugiyama, Toyomi Meguro, and Yasuhiro Minami.
\newblock Preference-learning based inverse reinforcement learning for dialog
  control.
\newblock In \emph{Thirteenth Annual Conference of the International Speech
  Communication Association}, 2012.

\bibitem[Taylor et~al.(2011)Taylor, Suay, and Chernova]{taylor2011integrating}
Matthew~E Taylor, Halit~Bener Suay, and Sonia Chernova.
\newblock Integrating reinforcement learning with human demonstrations of
  varying ability.
\newblock In \emph{The 10th International Conference on Autonomous Agents and
  Multiagent Systems-Volume 2}, pages 617--624. International Foundation for
  Autonomous Agents and Multiagent Systems, 2011.

\bibitem[Todorov et~al.(2012)Todorov, Erez, and Tassa]{todorov2012mujoco}
Emanuel Todorov, Tom Erez, and Yuval Tassa.
\newblock Mujoco: A physics engine for model-based control.
\newblock In \emph{Intelligent Robots and Systems (IROS), 2012 IEEE/RSJ
  International Conference on}, pages 5026--5033. IEEE, 2012.

\bibitem[Wilson et~al.(2012)Wilson, Fern, and Tadepalli]{wilson2012bayesian}
Aaron Wilson, Alan Fern, and Prasad Tadepalli.
\newblock A bayesian approach for policy learning from trajectory preference
  queries.
\newblock In \emph{Advances in neural information processing systems}, pages
  1133--1141, 2012.

\bibitem[Wise et~al.(2016)Wise, Ferguson, King, Diehr, and
  Dymesich]{wise2016fetch}
Melonee Wise, Michael Ferguson, Derek King, Eric Diehr, and David Dymesich.
\newblock Fetch and freight: Standard platforms for service robot applications.
\newblock In \emph{Workshop on Autonomous Mobile Service Robots}, 2016.

\bibitem[Ziebart et~al.(2008)Ziebart, Maas, Bagnell, and
  Dey]{ziebart2008maximum}
Brian~D Ziebart, Andrew~L Maas, J~Andrew Bagnell, and Anind~K Dey.
\newblock Maximum entropy inverse reinforcement learning.
\newblock In \emph{AAAI}, volume~8, pages 1433--1438. Chicago, IL, USA, 2008.

\end{thebibliography}
\clearpage
\appendix
This appendix complements the RSS 2019 paper, ``Learning Reward Functions by Integrating Human Demonstrations and Preferences".
\subsection{Supplemental Videos}
We have provided the following videos in the ``Videos" folder to supplement our work:
\begin{enumerate}
    \item $\texttt{Demo.mov}$: This video shows a user teleoperating the robot using the keyboard interface.
    \item $\texttt{PrefT1.mov}$ and $\texttt{PrefT2.mov}$: These two videos show a preference query (two trajectories) generated by our system. Note that this pair of trajectories is clearly querying the user for whether she wants the robot arm to move towards the goal or away from the goal. Additionally, note the jaggedness of the trajectory: this is due to the highly non-convex nature of the optimization problem \eqref{obj_rank}.
    \item $\texttt{RolloutDemPref.mov}$: This video shows a sample trajectory generated by PPO, according to the reward function learned by DemPref (from a specific user). (In reality, the robot arm does get fairly close to the goal; we intentionally kept the table much lower when rolling-out behavior on the real robot to prevent collisions between the robot and the table. Users in our study were informed of this.) 
    \item $\texttt{RolloutIRL.mov}$: This video shows a sample trajectory generated by PPO, according to the reward function learned by IRL (from the same user as above). Note the extremely poor performance of the robot -- this is discussed in Section \ref{user_study}.
\end{enumerate}

\subsection{Code}

The repository for this project is provided at the following link \url{https://github.com/malayandi/DemPrefCode}.

\prg{Dependencies.} To play around with our code, it will be easiest to use a Conda environment on an OSX system. Simply install Anaconda and then run the following commands (in the given order) from the $\texttt{DemPrefCode}$ directory to setup and activate the environment.
~\\
~\\
$\texttt{conda create --name dempref --file requirements.txt}$
$\texttt{source activate dempref}$
~\\

\prg{Instructions to Re-Run Experiments} To re-run any simulation experiment, run the following two commands from the DemPref directory, in the given order:
~\\
~\\
$\texttt{cd experiments/\{experiment\_name\}\_experiment}$
$\texttt{python \{experiment\_name\}\_experiment.py \{domain\_name\}}$
~\\
where $\texttt{\{experiment\_name\}}$ is one of ``main", ``update\_func", ``iterated\_corr" and $\texttt{\{domain\_name\}}$ is one of ``driver", ``lander", "fetch\_reach".

\prg{Raw Data and Processing Files.} For each of the simulation experiments, the raw data and Jupyter notebooks used to plot the figures seen in this paper have been provided. They can be found in the relevant directory for each experiments. (Enter the $\texttt{experiments}$ directory and you will see directories for the three simulation experiments.)

\subsection{Additional Experimental Details}

\prg{System Used to Run Experiments.} Our system had 8 cores and 4 GB of memory.

\prg{Hyperparameters.} The hyperparameters used were, unless otherwise noted, constant across all experiments (including the user study). Minimal hyperparameter tuning was performed. The hyperparameters are as follows:
\begin{enumerate}
    \item $\beta^D$: 0.1
    \item $\beta^R$: 5
    \item Number of samples used in Monte Carlo approximation to objective in \eqref{obj_rank}: 50,000
\end{enumerate}

\prg{True Reward Function.} As discussed in Section \ref{domains}, we chose a ``true" weight vector for each domain to use in our simulation experiments. We chose a weight vector that seemed reasonable in each domain. No tuning was performed. The weight vector for each domain is as follows:
\begin{enumerate}
    \item Driver: \texttt{[0.5, -0.2, 0.2, -0.7]}
    \item Lunar Lander: \texttt{[-0.4, 0.4, -0.2, -0.7]}
    \item Fetch Reach: \texttt{[-0.6, -0.3, 0.9]}
\end{enumerate}

Any further experimental details not found here can be found in the provided code.

\end{document}